\documentclass[pdflatex,sn-mathphys-num]{sn-jnl}

\usepackage{graphicx}%
\usepackage{multirow}%
\usepackage{amsmath,amssymb,amsfonts}%
\usepackage{amsthm}%
\usepackage{mathrsfs}%
\usepackage{array}%
\usepackage{xspace}%
\usepackage{booktabs}
\usepackage[title]{appendix}%
\usepackage{xcolor}%
\usepackage{textcomp}%
\usepackage{manyfoot}%
\usepackage{booktabs}%
\usepackage{algorithm}%
\usepackage{algorithmicx}%
\usepackage{algpseudocode}%
\usepackage{listings}%

\theoremstyle{thmstyletwo}%

\theoremstyle{thmstylethree}%

\begin{document}

\title[Article Title.]{Assessment of Transformer-Based Encoder-Decoder Model for Human-Like Summarization.}

\author*[1]{\fnm{Sindhu} \sur{ Nair}}\email{sindhu.nair@djsce.ac.in}

\author[2]{\fnm{Dr. Y.S.} \sur{Rao}}\email{ysrao@spit.ac.in}
\equalcont{These authors contributed equally to this work.}

\author[3]{\fnm{Dr. Radha} \sur{Shankarmani}}\email{rsmadam@gmail.com}
\equalcont{These authors contributed equally to this work.}

\affil*[1]{\orgdiv{Asst. Professor}, \orgname{D.J. Sanghvi College of Engineering}, \orgaddress{\street{JVPD}, \city{Mumbai}, \postcode{400 056}, \state{Maharashtra}, \country{India}}}

\affil[2,3]{\orgdiv{Professor}, \orgname{Sardar Patel Institute of Technology}, \orgaddress{\street{Andheri (W)}, \city{Mumbai}, \postcode{400 048}, \state{Maharashtra}, \country{India}}}


\abstract{In recent times, extracting valuable information
from large text is making significant progress. Especially in
the current era of social media, people expect quick bites of
information. Automatic text summarization seeks to tackle this
by slimming large texts down into more manageable summaries.
This important research area can aid in decision-making by
digging out salient content from large text. With the progress
in deep learning models, significant work in language models
has emerged. The encoder-decoder framework in deep learning
has become the central approach for automatic text summarization.
This work leverages transformer-based BART model
for human-like summarization which is an open-ended problem
with many challenges. On training and finetuning the encoder-decoder
model, it is tested with diverse sample articles and
the quality of summaries of diverse samples is assessed based
on human evaluation parameters. Further, the finetuned model
performance is compared with the baseline pretrained model
based on evaluation metrics like ROUGE score and BERTScore. Additionally, domain adaptation of the model is required for
improved performance of abstractive summarization of dialogues
between interlocutors. On investigating, the above popular evaluation metrics are found to be insensitive to factual errors. Further investigation of the summaries generated by finetuned model is done using the contemporary evaluation metrics of factual consistency like WeCheck and SummaC. Empirical results on BBC News articles highlight that the gold standard summaries written by humans are more factually consistent by 17\%\ than the abstractive summaries generated by finetuned model.}

\keywords{Automatic Text Summarization, Deep learning,
Transformers, BART-LARGE-CNN, ROUGE score, BERTScore, WeCheck, SummaC.}



\maketitle

\section{Introduction}\label{sec1}

Enormous amount of data is available in unstructured format in the internet age as compared to structured data. Such data is generated on a daily basis like online news articles, research papers, e-mail messages, e-books to name a few. 
With the rapid rise in web text data, extracting semantic value from long text is a vitally important research. The sequence-to-sequence(seq2seq) framework in deep learning has become the predominant approach for automatic text summarization. The end-to-end training in seq2seq framework
learns the semantic mapping between the source documents and its corresponding summaries. 
Despite huge advances in automatic text summarization models, it is still challenging to generate abstractive summaries of good quality.

Humans summarize written text by first understanding the content of the document. The next step is to identify most important or salient information. Finally, this information is reworded in a compressed form. Hence, for a synopsis identify important information, delete non-essential extraneous information and then, rewrite the remaining information to make it more general and more compact. It is important to comprehend the content of the document to get the central theme and summarize it. 

There exist two overarching approaches for text summarization. Extractive summarization
directly copies content straight from the input text to create synopsis. The model may copy whole sentences or copy words/phrases. One can think of it as using a highlighter to point out the important parts of the document. This is the approach employed by many classical summarization works, especially before the advent of neural networks. Abstractive summarization creates a summary without being limited to only words and sentences within the source document. Often this is done by generating a summary one word at a time by picking a word from a set vocabulary, until a whole summary has been created. One can think of it as how a human might write a summary in their own words. This method allows for more compression since lengthy sentences can be reworded into simpler expressions [1]. Neural abstractive summarization has reached novel heights, especially with the progress in deep learning models. End-to-end abstractive summarization models must perform two tasks implicitly at the same time. 1) Content selection: salient sentences or words from the source text must be selected. 2) Surface realization: a summary must be generated which successfully merges the selected content together [1]. 

This paper addresses the following questions. Q1: Does the baseline pre-trained BART-LARGE-CNN model when finetuned on a small dataset boost the abstractive summarization performance? Q2: If the input is much diverse from the training dataset say dialogues between interloculators, will domain adaptation give the enhanced performance on abstractive summarization? 
Q3: Is there an alignment or correlation between the popular evaluation metrics like ROUGE score, BERTScore and the human evaluation of the synopsis generated of diverse articles? Do the existing evaluation metrics capture the factual errors? If not,
How is the factual consistency of the generated synopsis measured and is it correlated to the human evaluation? Q4: Are the summaries generated by the finetuned model factually consistent in comparison to the corresponding gold standard summaries written by humans?

\section{Conceptual Framework For Abstractive Summarization}

The abstractive summarization method produces concise summaries with novel words which incorporates both syntax as well as semantics. The parameters to check quality of summary include fluency, non-repetition, meaningfulness, flow, grammar and syntax, semantics, conciseness, saliency, information diversity, information coverage and accuracy. Deep neural encoder-decoder models are predominantly applied for various NLP tasks including summarization [1].

\begin{figure}[htbp]
    \centering
    \centerline{\includegraphics[width=9.6cm]{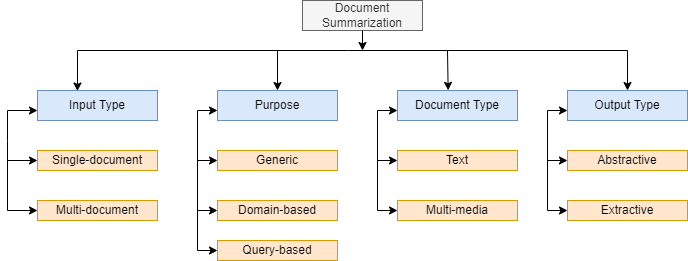}}
    \caption{Document Summarization Techniques.}
    \label{fig}
\end{figure}

Fig. 1 depicts the categories of document summarization techniques which can span from single documents to multiple documents [2] including only text or text augmented with multimedia content like images, audio and video. The objective of summarization spans from generic to query-specific. 
The output of the task is purely extractive if it extracts salient phrases, sentences from the source to create the abridged version. Alternatively, with the introduction of novel words and re-phrasing, the approach is more human-like or abstractive.

\begin{figure}[htbp]
    \centering
    \centerline{\includegraphics[width=9.6cm]{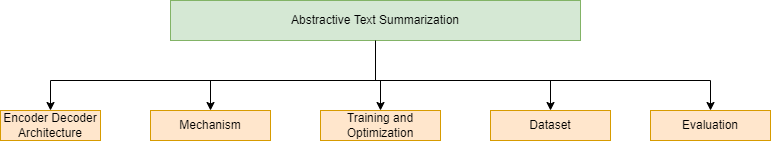}}
    \caption{The Conceptual Framework For Abstractive Human-like Summarization.}
    \label{fig}
\end{figure}

Fig. 2 depicts the five key elements of the conceptual framework for abstractive human-like summarization [2]. The principal component is the encoder-decoder framework which is further finetuned on custom dataset with specific training strategies. Standard metrics like ROUGE score and BertScore are used for evaluating the performance of the encoder-decoder models.

\subsection{Encoder-Decoder Model}

The training mechanism used to map sequential data from one domain to another is seq2seq learning. The mapping is based on embedding layer and text generation techniques. Such a framework consists of encoder and decoder. In the encoding stage,
the model examines the input and maps it into an intermediate embedding layer. In the 
decoding stage, from the intermediate vector mapping, the output sequence is predicted.
Additionally, the basic model can be enriched with a feature-rich encoder [2]. 

\begin{figure}[htbp]
    \centerline{\includegraphics[width=9.6 cm]{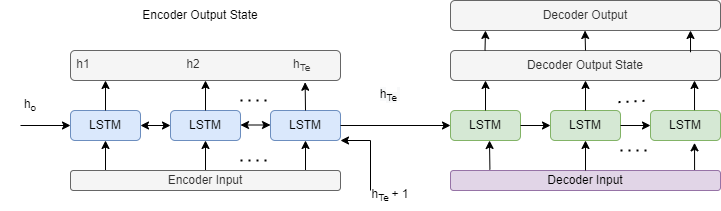}}
    \caption{Simple seq2seq model.}
    \label{fig}
\end{figure}

Fig. 3 depicts the structure of simple seq2seq model based on LSTM in the pre-transformer era. In the pre-transformer era, the encoder-decoder architecture includes  RNN, LSTM, GRU adept for processing sequential data. However, they fail to handle long range dependencies. Further, due to sequential processing of input, the training time and inference time is longer while handling long range dependencies. 

\begin{figure}[htbp]
    \centerline{\includegraphics[width=9.6cm]{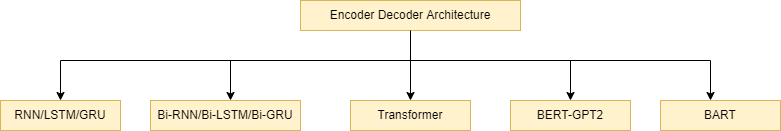}}
    \caption{Encoder Decoder Framework.}
    \label{fig}
\end{figure}

Fig. 4 depicts various combinations for Encoder-decoder Framework [2]. RNN, LSTM, GRU combinations for encoder-decoder framework was surpassed by the advent of the transformer era leading to the BART model with BERT-like encoder and GPT-like autoregressive decoder. The limitations of pre-transformer era are overcome due to the inherent parallel processing and self-attention mechanism in transformers [3]. The Transformer model is a neural network architecture introduced by Google researchers in 2017, focusing on an attention-based mechanism to improve natural language processing tasks [4]. The scalability and training speed is enhanced in transformers due to parallel processing. The attention mechanism improves the model’s ability to capture context and relationships within the data. This work leverages the transformer-based BART model for abstractive summarization. 

\begin{figure}[htbp]
    \centerline{\includegraphics[width=4.0cm]{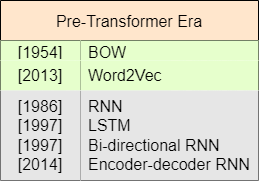}}
    \caption{Pre-Transformer Timeline of NLP.}
    \label{fig}
\end{figure}

Fig. 5 depicts the timeline of NLP from BOW up to the attention-based Transformers in 2017. In the pre-transformers era, the Bag of Words scores were improved in the TF-IDF where frequently occurring words were assigned low scores. This was further replaced by word embeddings like word2vec, glove, fasttext. Recurrent Neural Networks was more contextual with further advances like Bidirectional RNNs and encoder-decoder RNNs. LSTMs were further preferred for sequential input to capture long range dependencies [3].

\begin{figure}[htbp]
    \centerline{\includegraphics[width=6.6cm]{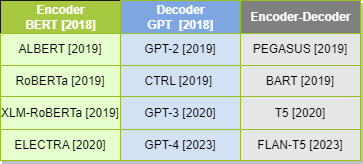}}
    \caption{Transformer Family With Timeline.}
    \label{fig}
\end{figure}

Fig. 6 gives a glimpse of the transformer family alongwith the timeline. The attention mechanism in transformers computes contextual word embeddings and is able to capture the semantic alignment between the input and the output [4]. The encoder only models include BERT [5] and its variations like ALBERT, RoBERTa, DistilBERT, XLM-RoBERTa, ELECTRA. The decoder only models its GPT and its improvements. The encoder-decoder transformer models include pegasus, BART, T5.

\subsection{Context-dependent Transformer-based Models}

Context-dependent approach output different word embeddings for the input word on the basis of the neighbouring words or the semantics of usage of the word. BERT, XLM, RoBERTa, ALBERT are some of the transformer based techniques for context-dependent approach. They are trained on large corpora, which can be cross-domain. The training has variations in key hyperparameters, mini-batches, learning rates.

Transformers are just another type of neural network which only uses feedforward layers and attention layers [4]. The Transformer includes an encoder and a decoder, each made up of layers containing a multi-head self-attention mechanism and a feed-forward neural network. The encoder is responsible to give the context vector that contains information of the input sequences. Such a vector gives the summary of the input. The objective to form a context vector is to get a numerical representation such that ideally there is no loss of information. Decoder generates the output sequence with the help of the context vector [4]. The attention mechanism in transformers ranges from self-attention in the encoder as well as the decoder to the cross-attention from the encoder to the decoder. 

The input is vectorized in the embedding layer. Positional Encoding layer tracks the relative position of words such that the relationship between words and the overall context and semantics is better captured. The number of encoders in the Encoder Layer can vary from 6 to 12 depending on the complexity of the transformer model. The encoders are stacked one over the other in the Encoder layer. The same pattern is duplicated in the decoder stack. Each encoder/decoder layer has self-attention which is looped several times which is called multi-head attention. This iterative multi-head attention teamed up with the feed-forward layer makes the transformer very powerful. It is able to process long sequences of text and capture the context as well as the semantics of the text in a very effective manner [4]. The decoder output is fed to a softmax layer through a linear layer for a probability distribution.

\begin{figure}[htbp]
    \centerline{\includegraphics[width=9.6cm]{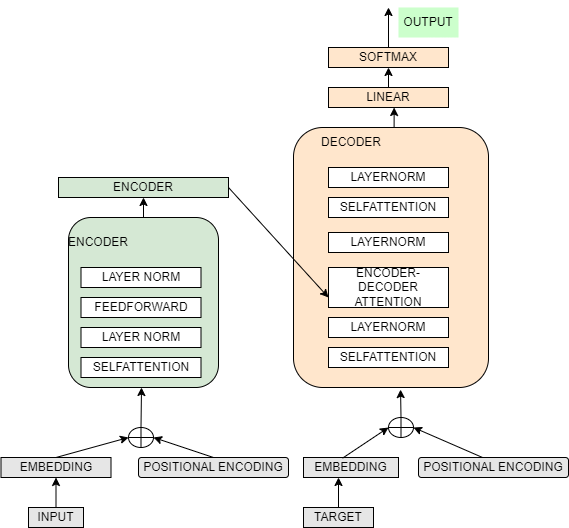}}
    \caption{Main Components Of The Transformer Architecture.}
    \label{fig}
\end{figure}

Fig. 7 depicts the in-depth architecture of the encoder-decoder transformer.
There are numerous variations to the original transformer. Depending on the task for supervising learning of text ranging from text classification, sentiment analysis, summarization, translation to question answering, the model could have either only encoder, or only decoder. For some tasks such as translation and summarization, the encoder-decoder model is preferred. BART is one such encoder-decoder sequence to sequence model which is also applied for comprehension tasks [6]. One example is Bart-Large-CNN model wherein the base BART model is trained on CNN Daily Mail which is a huge dataset of text-summary pairings. A very popular transformer model which has only encoder is BERT [5]. It is commonly used for text classification and sentiment analysis. GPT is another transformer model which is having only decoder and popularly used for text generation.

BART is a type of transformer which has a bidirectional encoder and an autoregressive decoder. Hence, justifiably BART stands for Bidirectional and Auto Regressive Transformers. This sequence-to-sequence model has BERT-like encoder and decoder-like GPT. Depending on the number of layers in the encoder stack as well as the decoder stack, the complexity of the model can be varied. Hence, the base model can range from 6 layers to 12 layers in the encoder/decoder stack. Also, the huge labelled dataset for pre-training, can lead to multiple versions of the model ranging from BART-LARGE-CNN to BART-SAMSUM. The pretraining objective is denoising where the original sentences are re-shuffled alongwith replacement of text spans with a single mask token [6].

\section{Related Work} The recent trend is the use of deep neural networks for the task of summarization which is abstractive. Basic deep sequence-to-sequence models have the problem of long-term dependencies in long sequences. Convolutional neural networks (CNNs) have met great success in abstractive summarization [7], [8] but fail to tackle long documents. One approach is to augment attention mechanism which would enable the encoder to focus only on the most salient parts of text. This is an efficient way of memory usage. The encoder-decoder attention-based transformer models like BART [6] from Facebook AI, Pegasus [9] and T5 [10] from Google are the state-of-the-art models for abstractive summarization. Another approach is to leverage Pointer Generator networks which is a solution to the problem of inaccurate factual details and out-of-vocabulary words problem [11].As discussed in [12], the encoder-decoder model is based on a double attention pointer network (DAPT). In DAPT, the self-attention mechanism make the model apt for long-term dependencies in long sequences and the pointer network generates accurate summaries. In [12], the coverage mechanism is augmented over the double attention pointer network to avoid duplication by keeping track of what has been generated in the summary. Generative adversarial networks [13] have been used in many applications with promising results. Previous research has shown the effectiveness of generative adversarial networks in text summarization. Researchers in [14] propose a novel Hierarchical Human-like deep neural network for ATS (HH-ATS), inspired by the process of how humans comprehend an article and write the corresponding summary. Specifically, this hybrid model consists of dual discriminator generative adversarial network, attention-based knowledge-enhanced module and a multitask learning module. The human reading cognition consists of rough reading, active reading and postediting. The novel Hierarchical Human-like deep neural network for ATS (HH-ATS) apes the three phases of human reading cognition. Sequence-to-Sequence models with various deeplearning-based mechanisms like attention, coverage still suffer from exposure-bias problem, loss/evaluation mismatch, and lack of generalization. These problems are solved by leveraging Reinforcement Learning in [15]-[17]. Despite all the advances in deep learning and the recent trends in NLP task, assessment of the summaries based on human evaluation parameters like factual consistency and faithfulness of the generated summary to the source text is a challenging task. 

This work investigates the performance of the BART model for human-like abstractive summarization which is a gap in related work. It focusses on training and fine-tuning the BART model on BBC News Dataset, to generate concise summaries. The evaluation metrics used include ROUGE Score, BERTScore, and FactCC, to measure the quality and factual consistency of the generated summaries. 

\section{Experimental Setup}

\subsection{Dataset}\label{AA}

The transformer-based BART-LARGE-CNN model is pre-trained on huge corpus of CNN/DailyMail Dataset. It consists of more than 300k unique English news articles penned by authors at CNN and the Daily Mail. This dataset includes article-highlights pairings. The dataset is further split into huge amount of training data and equivalent portion of validation and test data.
Post pre-training, the BART-LARGE-CNN model is finetuned with BBC News Dataset. It comprises of BBC News articles which are classified as business articles, entertainment articles, political articles, sports articles and technical articles. Each category consists of article-summary pairings. 

The dataset is available at  https://www.kaggle.com/antoniobap/datamining-bbc-news/data. Table I illustrates the dataset statistics of BBC News Dataset. This dataset has news documents with it's corresponding synopsis. Table II illustrates the categorization of the news articles into business articles, entertainment articles, political articles, sports articles and technical articles. The dataset has a fair distribution of the different categories of news articles. Thus, there is a good balance in the categorization of documents.

\begin{table*}[h]
\caption{Dataset Statistics of BBC News}\label{tab1}%
\begin{tabular}{@{}lll @{}}
\toprule
TrainData & ValidationData & TestData \\
\midrule
2225    & 998   & 998 \\
\midrule
\end{tabular}
\end{table*}

\begin{table*}[h]
\caption{BBC NEWS Dataset Description Of 2225 Documents}\label{t4}%
\begin{tabular}{@{}ll @{}}
\toprule
TOPICAL AREAS/ Category & FILES count\\
\midrule
Business    & 510 \\
Entertainment    & 386 \\
Politics    & 417 \\
Sports    & 511 \\
Technical    & 401 \\
\midrule
\end{tabular}
\end{table*}

\subsection{Abstractive summarization with Huggingface}\label{AA}

Hugging Face Transformers library uses a pre-trained transformer model BART with it's large-sized version, fine-tuned on CNN Daily Mail powered by Meta (Facebook) to generate abstractive summaries from input text. BART-LARGE-CNN uses BartTokenizer and generate() methods used for conditional generation tasks like summarization.

\begin{algorithm}[!ht]
\caption{Fine Tuning and Model Training}\label{<alg1>}
Input : Parameters, Pre-trained Model\\
Dataset : BBC news dataset\\
Output : Trained model\\
Steps :\\
1. Import the libraries, dependencies.\\
2. Import and preprocess the input data.\\
3. Import the model.\\
4. Train the model by fine tuning on new data.\\
5. Generate Predictions and inference.\\
\end{algorithm}

The maximum length is set for the output and input sequences. Finally, the predictions are generated and the results of the trained model are evaluated.

The performance metrics are evaluated for the finetuned BART-LARGE-CNN model. The finetuned BART-LARGE-CNN model is tested with samples and the generated summaries are evaluated. 

The quality of the summary generated is evaluated with ROUGE score [18] as well as BERTScore [19].

\section{Empirical Results}

The `facebook/bart-large-cnn' pre-trained model from huggingface is used in the experiments.  
The Bart tokenizer is used for the encoding which takes into consideration the position of the words in the sentence. The Bart tokenizer is built from the GPT-2 tokenizer.
The summaries generated were fluent. Hence, the summaries generated by the BART pre-trained model demonstrating the effectiveness of the BART model for text understanding. There was significant improvement in the ROUGE score after fine-tuning the pre-trained BART-LARGE-CNN model with the BBC News Dataset. 
The ROUGE scores for finetuned BART-LARGE-CNN during training for 10 epochs are compiled in Table \ref{t1}.

\begin{table*}[h]
\caption{Fine Tuning Results of ROUGE score Upto 10 Epochs of Bart-Large-CNN.}\label{t1}%
\begin{tabular}{@{}llll @{}}
\toprule
Epoch & ROUGE1 & ROUGE2 & ROUGEL \\
\midrule
0 & 0.548 & 0.426 & 0.365 \\
1 & 0.537 & 0.399 & 0.381 \\
2 & 0.577 & 0.437 & 0.395 \\
3 & 0.550 & 0.412 & 0.389 \\
4 & 0.577 & 0.430 & 0.414 \\
5 & 0.554 & 0.415 & 0.389 \\
6 & 0.602 & 0.467 & 0.430 \\
7 & 0.603 & 0.471 & 0.414 \\
8 & 0.592 & 0.458 & 0.402 \\
9 & 0.604 & 0.470 & 0.412 \\
\midrule
\end{tabular}
\end{table*}

\begin{figure*}[htbp]
    \includegraphics[width=8.6cm]{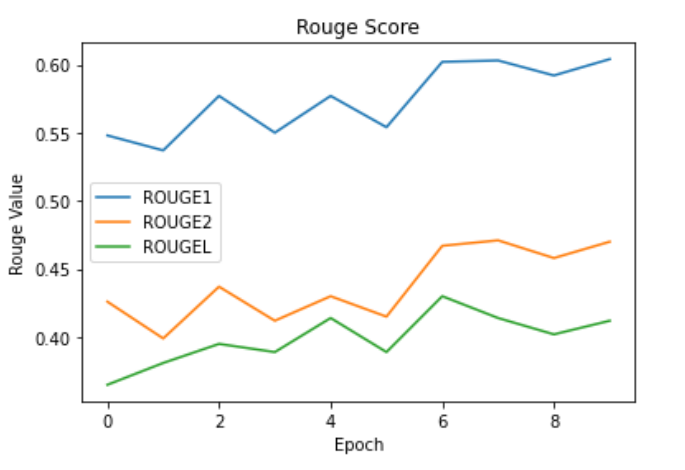}
    \caption{ROUGE score of Bart-Large-CNN.}
    \label{fig}
\end{figure*}

\begin{table*}[h]
\caption{Fine Tuning Results of BERTScore Upto 10 Epochs of Bart-Large-CNN.}\label{t2}%
\begin{tabular}{@{}llll @{}}
\toprule
Epoch & BERTScoreP & BERTScoreR & BERTScoreF1 \\
\midrule
0 & 0.912 & 0.889 & 0.900 \\
1 & 0.913 & 0.886 & 0.899 \\
2 & 0.912 & 0.892 & 0.902 \\
3 & 0.910 & 0.890 & 0.899 \\
4 & 0.911 & 0.896 & 0.904 \\
5 & 0.910 & 0.891 & 0.900 \\
6 & 0.916 & 0.901 & 0.908 \\
7 & 0.916 & 0.901 & 0.908 \\
8 & 0.913 & 0.899 & 0.905 \\
9 & 0.914 & 0.900 & 0.907 \\
\midrule
\end{tabular}
\end{table*}

Table \ref{t2} shows the results of the BERTScore of fine-tuned BART-LARGE-CNN model during training for 10 epochs. The ROUGE score and BERTScore obtained for each epoch while training is plotted in Fig. 8 and Fig. 9.

\begin{figure}[htbp]
    \includegraphics[width=8.6cm]{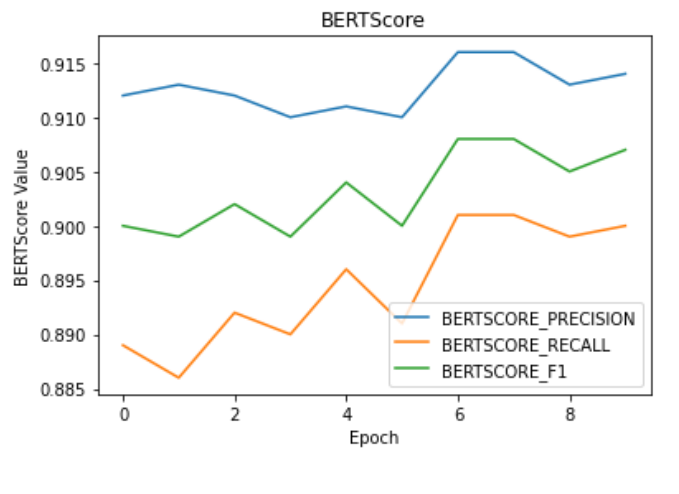}
    \caption{BERTScore of Bart-Large-CNN.}
    \label{fig}
\end{figure}

The input samples for testing are compiled in Table \ref{t4}. It includes Technical Article, Article with many numeric facts, Article with flow (cause, effect, solution), Sports News Article, Health News Article and Political News Article.

\begin{table*}[h]
\caption{Input Samples For Testing.}\label{t4}%
\begin{tabular}{@{}ll @{}}
\toprule
Input & Type\\
\midrule
Sample1 & Technical Article\\
Sample2 & Article with many numeric facts\\
Sample3 & Article with flow (cause, effect, solution)\\
Sample4 & Sports News Article\\
Sample5 & Health News Article\\
Sample6 & Political News Article\\
Sample7 & Dialogue between two interlocutors\\
\midrule
\end{tabular}
\end{table*}

For evaluation of the summaries, a popular metric is ROUGE score. It matches the words/tokens between the predicted summary and the reference summary. The reference summary is the summary manually written by a human and hence is coherent, salient, accurate without deviating from the source text. It captures the semantics of the document. ROUGE score [18] has many variations ranging from ROUGE1, ROUGE2, ROUGEL to ROUGELsum. ROUGE1 indicates 1-gram overlap between the predicted summary and the reference summary while ROUGE2 indicates 2-gram overlap between the two. ROUGEL is the overlap of the longest subsequence between the generated summary and the reference summary. However, since ROUGE score follows n-gram matching or overlap, it falls short of capturing the semantics of the text. Another evaluation metric which captures the semantics and the similarity between the generated summary and the reference summary is BERTScore [19]. It can be further split into precision, recall and F1-score. Hence, BERTScore is more contextual and hence is an improvised evaluation metric. In addition to the evaluation metrics ROUGE score and BERTScore, the human evaluation provides further validation. 
\begin{table*}[h]
\caption{Human Evaluation Parameters.}\label{t7}%
\begin{tabular}{@{}ll @{}}
\toprule
Parameter & Meaning\\
\midrule
Coherence  & Logical/clear\\
Factual Consistency/Accuracy &  Correctness \\
Duplication  &   Redundancy/Repetition\\
Saliency &  Main points coverage \\
Fluency & Grammatical Correctness\\
Faithfulness  &  Deviation From source \\

\midrule
\end{tabular}
\end{table*}
Table \ref{t7} lists the human evaluation parameters for assessment of the generated summaries by the deep learning model. The human evaluation based on the parameters specified in Table \ref{t7} evaluates the quality of the summaries generated. The human evaluation parameters range from coherence, saliency, fluency, faithfulness to accuracy. 
The fine-tuned BART-LARGE-CNN model is evaluated on the testing samples. The predicted summaries are used to calculate the ROUGE score [18] and the BERTScore [19]. 

\begin{table*}[h]
\caption{ROUGE score and BERTScore for Sample1.}\label{t8}%
\begin{tabular}{@{}llll @{}}
\toprule
Model & ROUGE1 & ROUGE2 & ROUGEL \\
\midrule
Pretrained & 0.47  & 0.44  & 0.46  \\
Finetuned & 0.51  & 0.46  & 0.49  \\
\toprule
Model & BERTScorePrecision & BERTScoreRecall & BERTScoreF1-score \\
\midrule
Pretrained & 0.94  & 0.87  & 0.90  \\
Finetuned & 0.95  & 0.88  & 0.91 \\
\midrule
\end{tabular}
\end{table*}

Table \ref{t8} compares the performance metrics between the pretrained model and the finetuned model for Sample1 clearly showing an improvement in the quality of summaries generated by the finetuned model.

\begin{table*}[h]
\caption{ROUGE score and BERTScore for Sample2.}\label{t9}%
\begin{tabular}{@{}llll @{}}
\toprule
Model & ROUGE1 & ROUGE2 & ROUGEL \\
\midrule
Pretrained & 0.21
 & 0.05
 & 0.14
 \\
Finetuned & 0.59
 & 0.46
 & 0.51
 \\
\toprule
Model & BERTScorePrecision & BERTScoreRecall & BERTScoreF1-score \\
\midrule
Pretrained & 0.86
 & 0.84
 & 0.85
 \\
Finetuned& 0.92
 & 0.92
 & 0.92
 \\
\midrule
\end{tabular}
\end{table*}

Table \ref{t9} compares the performance metrics between the pretrained model and the finetuned model for Sample2 which is an article with many numeric facts. The results show a much improved performance with the finetuned model.

\begin{table*}[h]
\caption{ROUGE score and BERTScore for Sample3.}\label{t10}%
\begin{tabular}{@{}llll @{}}
\toprule
Model & ROUGE1 & ROUGE2 & ROUGEL \\
\midrule
Pretrained  & 0.62
 & 0.55
 & 0.55
 \\
Finetuned & 0.62
 & 0.55
 & 0.54
 \\
\toprule
Model & BERTScorePrecision & BERTScoreRecall & BERTScoreF1-score \\
\midrule
Pretrained  & 0.95
 & 0.899
 & 0.92
 \\
Finetuned & 0.946
 & 0.899
 & 0.922
 \\
\midrule
\end{tabular}
\end{table*}

Table \ref{t10} compares the performance metrics between the pretrained model and the finetuned model for Sample3 which is an article with a flow. The finetuned results show no improvement when tested with Sample3 indicating that the BART model has to be finetuned further with articles with a reasoning.

\begin{table*}[h]
\caption{ROUGE score and BERTScore for Sample4.}\label{t11}%
\begin{tabular}{@{}llll @{}}
\toprule
Model & ROUGE1 & ROUGE2 & ROUGEL \\
\midrule
Pretrained  & 0.54
 & 0.38
 & 0.39
 \\
Finetuned & 0.61
 & 0.51
 & 0.50
 \\
\toprule
Model & BERTScorePrecision & BERTScoreRecall & BERTScoreF1-score \\
\midrule
Pretrained  & 0.93
 & 0.899
 & 0.92
 \\
Finetuned & 0.94
 & 0.91
 & 0.93
 \\
\midrule
\end{tabular}
\end{table*}

\begin{table*}[h]
\caption{ROUGE score and BERTScore for Sample5.}\label{t12}%
\begin{tabular}{@{}llll @{}}
\toprule
Model & ROUGE1 & ROUGE2 & ROUGEL \\
\midrule
Pretrained  & 0.40
 & 0.20
 & 0.30
 \\
Finetuned & 0.49
 & 0.26
 & 0.36
 \\
\toprule
Model & BERTScorePrecision & BERTScoreRecall & BERTScoreF1-score \\
\midrule
Pretrained  & 0.89
 & 0.87
 & 0.88
 \\
Finetuned & 0.89
 & 0.88
 & 0.88
 \\
\midrule
\end{tabular}
\end{table*}

\begin{table*}[h]
\caption{ROUGE score and BERTScore for Sample6.}\label{t13}%
\begin{tabular}{@{}llll @{}}
\toprule
Model & ROUGE1 & ROUGE2 & ROUGEL \\
\midrule
Pretrained  & 0.70
 & 0.61
 & 0.70
 \\
Finetuned & 0.74
 & 0.69
 & 0.74
 \\
\toprule
Model & BERTScorePrecision & BERTScoreRecall & BERTScoreF1-score \\
\midrule
Pretrained  & 0.96
 & 0.89
 & 0.92
 \\
Finetuned & 0.97
 & 0.92
 & 0.95
 \\
\midrule
\end{tabular}
\end{table*}

Table \ref{t11}, Table \ref{t12} and Table \ref{t13} show an improvement in the quality of summaries generated by the finetuned model for Sample4, Sample5 and Sample6 respectively.

\begin{table*}[h]
\caption{Fine Tuning Results of BART-LARGE-CNN on SAMSum DATASET.}\label{t15}%
\begin{tabular}{@{}lllll @{}}
\toprule
Epoch & ROUGE1 & ROUGE2 & ROUGEL & ROUGELsum\\
\midrule
0 & 48.24  & 25.93  & 41.11 & 44.77 \\
1 & 48.41  & 26.25  & 41.29 & 44.82 \\
2 & 48.88  & 26.95  & 41.67 & 45.38 \\
3 & 49.15  & 26.93  & 41.55 & 45.44 \\ 
\midrule
\end{tabular}
\end{table*}

\begin{table}[h]
\caption{ROUGE score and BERTScore for Sample7.}\label{t14}%
\begin{tabular}{@{}llll @{}}
\toprule
Model & ROUGE1 & ROUGE2 & ROUGEL \\
\midrule
Pretrained & 0.34
 & 0.07
 & 0.22
 \\
FinetunedOnBBC & 0.26
 & 0.06
 & 0.15
 \\
FinetunedOnSAMSUM & 0.74
 & 0.66
 & 0.72
 \\

\toprule
Model & BERTScorePrecision & BERTScoreRecall & BERTScoreF1-score \\
\midrule
Pretrained & 0.87
 & 0.87
 & 0.87
 \\
FinetunedOnBBC & 0.86
 & 0.85
 & 0.86
 \\
FinetunedOnSAMSUM  & 0.93
 & 0.96
 & 0.94
 \\
\midrule
\end{tabular}
\end{table}

Table \ref{t14}  compares the performance metrics between the pretrained model and the finetuned model for Sample7 which is a dialogue between two interlocutors. It was observed that the model finetuned on BBC News gave poor performance on testing with Sample7. Since Sample7 is a conversation between two people, domain adaptation was required by training the model on new dataset SAMSum. It is available on huggingface for research purposes which consists of conversations between two or more interlocutors and their respective summaries in third person. The empirical results on finetuning BART-LARGE-CNN on SAMSum dataset is displayed in Table \ref{t15}.

After the domain adaptation of the model with Samsum dataset, the model shows higher ROUGE scores and BERTScore on testing with Sample7 as visible in Table \ref{t14}.

\begin{table*}[h]
\caption{Human Evaluation of model generated synopsis.}\label{tt7}%
\begin{tabular}{@{}lllllll@{}}
\toprule
 Parameter & Coherence & Accuracy & Duplication & Saliency & Fluency & Faithfulness\\
\midrule
Sample1 & yes & yes & no & yes & yes & yes\\
Sample2 & yes & no & no &  yes & yes & no\\
Sample3 & yes & yes & no & yes & yes & no\\
Sample4 & yes & yes & no & no & yes & yes\\
Sample5 & yes & yes & no & yes & yes & yes\\
Sample6 & yes & yes & no & yes & yes & yes\\
Sample7 & yes & no & no & yes & yes & no\\
\midrule
\end{tabular}
\end{table*}
Table \ref{tt7} below depicts the human evaluation by 5 raters with Cohen-kappa-score of 0.75 for each sample manually evaluated. The human evaluators are briefed on the human evaluation parameters so that they can carry out the analysis of different samples.
Thus, the model-generated synopsis of above seven diverse articles is assessed, based on human evaluation parameters like coherence, factual consistency, duplication, saliency, fluency, faithfulness. The human score for each of the sample is evaluated by giving a score of 1 if the evaluation of 'yes' or 'no' matches the meaning of the parameter else it is given a score of 0. The average score is assigned as the human score for each of the sample.
Table \ref{ttt7} below depicts the comparison of popular metrics like ROUGEL, BERTScoreF1-score with the human score computed for the model generated synopsis of each of the seven samples.

\begin{table*}[h]
\caption{Comparison of popular metrics like ROUGEL, BERTScoreF1-score with the HumanScore of model generated synopsis.}\label{ttt7}%
\begin{tabular}{@{}lllllll @{}}
\toprule
 & ROUGEL & BERTScoreF1 & HumanScore\\
\midrule
Sample1 & 0.49 & 0.91 & 1\\
Sample2 & 0.51 & 0.92 & 0.67\\
Sample3 & 0.54 & 0.92 & 0.83\\
Sample4 & 0.5 & 0.93 & 0.83\\
Sample5 & 0.36 & 0.88 & 1\\
Sample6 & 0.74 & 0.95 & 1\\
Sample7 & 0.72 & 0.94 & 0.67\\
\midrule
\end{tabular}
\end{table*}

Table \ref{ttt8} below depicts the comparison of Factual Consistency Evaluation Metrics like WeCheck and SummaC with the HumanScore of model generated synopsis of each of the seven samples.

\begin{table*}[h]
\caption{Comparison of Factual Consistency Evaluation Metrics like WeCheck and SummaC with the HumanScore of model generated synopsis.}\label{ttt8}%
\begin{tabular}{@{}lllllll @{}}
\toprule
 & WeCheck & $SummaC_{zs}$ & $SummaC_{conv}$ & HumanScore\\
\midrule
Sample1 & 0.88 & 0.36 & 0.92 & 1\\
Sample2 & 0.32 & 0.31 & 0.82 & 0.67\\
Sample3 & 0.90 & 0.41 & 0.73 & 0.83\\
Sample4 & 0.89 & 0.20 & 0.21 & 0.83\\
Sample5 & 0.94 & 0.90 & 0.91 & 1\\
Sample6 & 0.85 & 0.90 & 0.91 & 1\\
Sample7 & 0.04 & -0.30 & 0.32 & 0.67\\
\midrule
\end{tabular}
\end{table*}
The correlation for 
ROUGEL, BERTScoreF1-score and HumanScore is displayed in Table \ref{tttt7} below which is based on Pearson correlation. As is evident in table below, the Pearson correlation between ROUGEL and the computed HumanScore is -0.26 while the correlation between BERTScoreF1-score and the computed HumanScore is -0.34.

As exhibited in Table \ref{tttt7} below, there is a strong positive correlation of the Factual Consistency Metrics like WeCheck and SummaC with the human score of the model generated summaries of the seven samples. The average factual consistency score by taking mean of WeCheck and SummaC score is found to have a pearson correlation value of 0.85 with the human score of the model generated summaries. The scipy package in python is used to determine that this strong and positive correlation is statistically significant with a p-value of 0.0149.

\begin{table*}[h]
\caption{The Pearson correlation of ROUGEL, BERTScoreF1, and Factual Consistency Metrics like WeCheck, SummaC with HumanScore.}\label{tttt7}%
\begin{tabular}{@{}lllllll @{}}
\toprule
 & ROUGEL & BERTScoreF1 & WeCheck & $SummaC_{zs}$ & $SummaC_{conv}$\\
\midrule
HumanScore &  -0.26 & -0.34 & 0.83 & 0.78 & 0.57\\

\midrule
\end{tabular}
\end{table*}
One of the key findings of this work was that existing evaluation metrics like ROUGE score and BERTScore fail to capture the factual inconsistency in the synopsis generated in terms of both factual accuracy and deviation from the source text. Though BERTScore is more semantic than ROUGE score which is based on n-gram overlap, it is still not sensitive to factual errors and fails to capture factual inconsistency in the synopsis generated. This has been validated by manual evaluation of the generated synopsis based on human evaluation parameters. Hence, contemporary metrics are investigated which capture the factual consistency of the text to check whether the synopsis generated is faithful to the source text.
\begin{table}[h!]
\centering
\caption{The average ROUGEL score and BERTScoreF1-score for 100 articles of BBC News Dataset with Average Length of Summaries Generated.}\label{ttttt7}
\begin{tabular}{|>{\centering\arraybackslash}m{2.5cm}|>{\centering\arraybackslash}m{1.1cm}|>
{\centering\arraybackslash}m{2.5cm}|>
{\centering\arraybackslash}m{3.5cm}|}
\hline
\textbf{\textit{Model}} & 
\textbf{\textit{Length}} & 
\textbf{\textit{ROUGEL}} & \textbf{\textit{BERTScoreF1-score}} \\ \hline
FinetunedBART & 91 & 0.33 & 0.88 \\ \hline

\end{tabular}
\end{table}

\begin{table}[t]
\centering
\caption{The Average Factual Consistency Metrics for model-generated synopsis of 100 articles of BBC News Dataset in comparison with the gold standard summaries written by humans. }\label{ttttt8}
\begin{tabular}{|c|c|c|c|c|>{\centering\arraybackslash}p{13mm}|>{\centering\arraybackslash}p{13mm}|>{\centering\arraybackslash}p{13mm}|>{\centering\arraybackslash}p{16mm}|>{\centering\arraybackslash}p{16mm}|>{\centering\arraybackslash}p{16mm}|>{\centering\arraybackslash}p{16mm}|}
\hline
\textit{\textbf{Model}}& \textit{\textbf{WeCheck}}& \textit{\textbf{$SummaC_{zs}$}} & \textit{\textbf{$SummaC_{conv}$}}\\
     \hline
  \textit{\textbf{FinetunedBART}} & 0.68 & 0.17  & 0.30   \\
  \hline
  \textit{\textbf{Gold standard Summary written by humans}} & 0.77 & 0.44  & 0.44 \\
  \hline
  
\hline
\end{tabular}
\end{table}
 The metrics analysed for factual consistency include WeCheck [22], and SummaC[23]. WeCheck [22] is a weakly supervised metric model trained with the real generated text rather than synthetic text. WeCheck framework consists of noise-aware fine-tuning and weak annotation. Thus, WeCheck is a factual consistency metric trained from weakly annotated samples. SummaC [23] is a NLI-based metric for summary inconsistency detection. It is further forked into the zero-short model and the model with the convolution layer. It combines sentence-level entailment scores for the final factual consistency score. Since SummaC [23] captures the factual consistency score at the sentence-level rather than document level, this contemporary metric has empirically proved to give best results for factual consistency score.

Further experiments are conducted with 100 articles of BBC News dataset described in Section 4.1. Table \ref{ttttt7} above gives the mean value of the ROUGE score and BERTScore for 100 BBC News Summaries generated by the finetuned BART model. The scores are computed by comparing the model generated summaries with the gold standard summaries written by humans. Table \ref{ttttt7} below also depicts the average length in tokens of summaries generated in comparison with human summaries.

Table \ref{ttttt8} above displays the average summarization performance metrics for factual consistency of model-generated synopsis of 100 articles of BBC News dataset in comparison with corresponding gold standard summaries written by humans. It depicts the mean value of the WeCheck score, $SummaC_{zs}$, and $SummaC_{conv}$ scores for 100 BBC News Summaries generated by the finetuned BART model.

\section{Results and Discussion}

Since the model is fine-tuned on BBC News Dataset, it performs well for newspaper articles. However, performs poorly when a dialogue is input to fine-tuned BART-LARGE-CNN. 
Hence, further experiments were done with diverse input samples ranging from sports news article, technical article, health article, political news article, article with many numerical facts, article with a flow starting with a problem and ending in a solution. It was observed that when these inputs were given to the fine-tuned BART-LARGE-CNN model, they generated good summaries which were fluent, free of grammatical errors, devoid of duplication and coherent. When compared with the human generated summary for above, they showed significant improvement over the baseline pre-trained model in terms of ROUGE score and BERTScore. This attempts to answer Q1 as the finetuned model shows enhanced abstractive summarization performance over the baseline pre-trained model. 

For Sample1 which is a technical article, Sample5 which is a health article and Sample6 which is a political news article, the summary generated by the finetuned model was fluent, free of grammatical errors, devoid of duplication, faithful, accurate, salient and coherent. This was based on the human evaluation as per the parameters listed in Table 6. However, for some samples, the generated summary had factual errors and deviated from the source text on human evaluation. For Sample2 which is an article with many numerical facts, the summary generated by the finetuned model had some factual errors and hence inaccurate. Therefore, it deviated from the source text.
For sample3 which is a article with a flow starting with a problem and ending in a solution, the summary generated by the finetuned model deviated from the source text and hence not faithful to the source text.
For sample4 which is a sports news article, the summary generated by the finetuned model did not include all the salient points of the sports news article. 

On testing with Sample7 which is a dialogue between two interlocutors, the evaluation metrics like ROUGE score and BERTScore showed a poor score in comparison to the baseline pretrained model. For domain adaptation, the pretrained model was finetuned with SAMSum dataset with consists of more than 16k tuples of conversations. Each tuple of the dataset contains conversations between two or more interlocutors and their corresponding summaries in third person. Further on testing this model finetuned on SAMSum dataset with Sample7, there was a much boosted performance with higher ROUGE score and BERTScore. However, the summary in third person was found to be fluent, free of grammatical errors, devoid of duplication, salient and coherent but with some factual errors and unfaithful on human evaluation. This attempts to answer Q2 where domain adaptation give the boosted performance on abstractive summarization for dialogues between interloculators as input.

The correlation measures indicate a low and negative correlation between the human score and popular metrics like ROUGEL and BERTScoreF1-score. This suggests that popular metrics like ROUGE and BERTScore are not in alignment with human evaluation and fails to capture factual consistency and faithfulness of the summary. 
This attempts to answer Q3 where in for some samples, inspite of enhanced ROUGE score and BERTScore, the synopsis deviates from the source text and contains factual errors. Hence, it is not in alignment with the human evaluation which identifies the factual inconsistency. Thus, it is confirmed that popular metrics of summarization like ROUGE score, BERTScore are insensitive to factual errors and any deviation from the source text. Hence, contemporary metrics for evaluating factual inconsistency in the synopsis like WeCheck and SummaC are investigated. They are found to have a strong and positive correlation with the human score of the model generated summaries of the seven samples. This correlation is found to be statistically significant.

The abstractive summaries of 100 BBC News articles generated by finetuned model are evaluated using contemporary metrics for factual consistency such as WeCheck and SummaC as well as popular metrics like ROUGE and BERTScore. The BBC News Dataset contains the gold standard summaries written by humans for the 100 BBC News articles. Using contemporary metrics like WeCheck and SummaC, the factual consistency score for the gold standard summaries written by humans are evaluated.
This attempts to answer Q4 as the empirical results highlight that abstractive summaries generated by the finetuned BART model are not as factually consistent as the gold standard summaries written by humans. This is exhibited in Table \ref{ttttt8} above based on the factual consistency metrics like Wecheck and SummaC.
In comparison with the gold standard summaries written by humans, there is a decrease of 9\%\ in the WeCheck score, 27\%\ in the $SummaC_{zs}$ score and 14\%\ in the $SummaC_{conv}$ score. Taking average of all the factual consistency metric scores, it is observed that there is a decrease of 17\%\ in the factual consistency evaluation metric of the BART generated abstractive summaries in comparison to the gold standard summaries written by humans.

The study addresses the challenges of factual inconsistency and domain adaptation, suggesting that existing evaluation metrics are inadequate in capturing factual errors and leveraging contemporary metrics for factual consistency. The study highlights that factual errors and deviations from the source text remain significant issues in the abstractive summaries generated by the finetuned deep learning model in comparison to the human summaries. The study addresses challenges related to factual inconsistency and domain adaptation. 

\section{Conclusion}

 Transformer-based deeplearning models are the SOTA for various NLP tasks. This paper investigates the performance of the SOTA encoder-decoder BART model for the summarization task. The empirical details shows good performance with finetuned approach as is evident by the ROUGE score and BERTScore.
 As against the pre-trained model, the finetuned model shows improved performance metrics with newspaper articles. However, domain adaptation was required for dialogues between interlocutors. However, some problems of factual inconsistency and deviation from the source still persisted inspite of good ROUGE score and BERTScore in case of some sample articles.  Human evaluation of the generated summaries were based on parameters such as coherence, factual consistency, duplication, saliency, fluency and faithfulness. Based on the human evaluation metrics, the quality of summaries was assessed. The summaries were found to be fluent, free of grammatical errors, devoid of duplication, salient and coherent. However, for some samples it deviated from the source text and hence not faithful to the source text. This factual inconsistency of the predicted summaries would hamper the usability of the deeplearning summarization systems especially in sensitive domains like medical, military operations, news media as it would be misleading with serious consequences. Design of innovative evaluation metrics which capture the factual accuracy and alignment of the generated synopsis with source text is an active research area since the existing evaluation metrics are insensitive to factual errors. Contemporary evaluation metrics for factual consistency like WeCheck and SummaC highlight the factual inconsistency in the abstractive summaries generated by the finetuned BART model.

The authors have no competing interests to declare that are relevant to the content of this article.

\section{Declarations}

Ethical Approval : not applicable.

Funding : No funding was received for conducting this study.

Availability of data and materials : The dataset is available at  \url{https://www.kaggle.com/antoniobap/datamining-bbc-news/data}

\end{document}